\documentclass{article}

\usepackage[accepted]{benelearn2014}
\usepackage{microtype}

\usepackage{graphicx} 

\usepackage{mlapa}
\usepackage{url}
\benelearntitlerunning{Anomaly Detection}

\begin{document}

\twocolumn[
\benelearntitle{Anomaly Detection Based on Aggregation of Indicators}

\benelearnauthor{Tsirizo  Rabenoro}{tsirizo.rabenoro@snecma.fr}
\benelearnauthor{J\'{e}r\^{o}me Lacaille}{jerome.lacaille@snecma.fr}
\benelearnaddress{ Health Monitoring Department, Snecma, Safran Group, Moissy Cramayel,
  France} 
\benelearnauthor{Marie Cottrell}{marie.cottrell@univ-paris1.fr}
\benelearnauthor{Fabrice Rossi}{fabrice.rossi@univ-paris1.fr}
\benelearnaddress{ SAMM (EA 4543),
  Universit\'{e} Paris 1, Paris, France}
\benelearnaddress{{\bf Keywords}: Anomaly Detection, Turbofan,	Health Monitoring 
          }
\vskip 0.3in
]

\begin{abstract}
Automatic anomaly detection is a major issue in various areas. Beyond mere
detection, the identification of the origin of the problem that produced the
anomaly is also essential. 

This paper introduces a general methodology that can assist human operators
who aim at classifying monitoring signals. 
The main idea is to leverage expert knowledge by generating a very
 large number of indicators. A feature selection method is used to keep only the most
  discriminant indicators which are used as inputs of a Naive Bayes
  classifier. The parameters of the classifier have been optimized indirectly by the
  selection process. Simulated data   designed to reproduce some of the anomaly types observed in real world
  engines. 
\end{abstract}

\section{Introduction}
Automatic anomaly detection is a major issue in numerous areas
and has generated a vast scientific literature
\cite{chandola2009anomaly}. 
Among the possible choices, statistical techniques
for anomaly detection are appealing because they can make use of expert knowledge
about the expected normal behaviour of the studied system. Thus they can
compensate for the limited availability of faulty observations (or more
generally of labelled observations). Those techniques are generally based on a
stationarity hypothesis. Numerous
parametric and nonparametric methods have been proposed to achieve this goal
\cite{basseville1995detection}.

However, statistical tests efficiency is highly dependent on the adequacy
between the assumed and actual data distribution.
 In addition, statistical methods rely on
meta-parameters, such as the length of the time window on which a change is
looked for. These meta-parameters have to be tuned to give maximal efficiency.

This article proposes to combine a (supervised) classification approach to
statistical techniques in order to obtain an automated anomaly detection
system that leverages both expert knowledge and labelled data sets. The main
idea consists in building a large number of binary
indicators that correspond to anomaly detection decisions taken by statistical
tests suggested by the experts, with varying (meta)-parameters. Then a feature
selection method is applied to the high dimensional binary vectors to select
the most discriminative ones, using a labelled data set. Finally, a classifier
is trained on the reduced binary vectors to provide automatic detection for
future samples. 
  
This approach has numerous advantages. 
On the classification point of view, it has been shown in
e.g. \cite{fleuret-2004} that selecting relevant binary features among a large
number of simple features can lead to very high classification accuracy in
complex tasks. In addition, using features designed by experts allows one to
at least partially interpret the way the classifier is making decisions as
none of the features will be of a black box nature. This is particularly
important in aircraft engine health monitoring context (see Section \ref{sec:context}). 
The indicators play also a homogenisation role by hiding the complexity of the
signals (in a way similar to the one used in
\cite{hegedus2011methodology}, for instance). On the statistical point of
view, the proposed approach brings a form of automated tuning: a test
recommended by an expert can be included in numerous variants. 
The feature selection process
keeps the most adapted parameters.

The rest of the paper is organized as follows. Section \ref{sec:context}
describes in more details Snecma's engine health monitoring context which
motivates this study.  Section \ref{sec:method} presents in more details the
proposed methodology. Section \ref{sec:results} presents the results obtained
on simulated data. 

\section{Application context}\label{sec:context}
\subsection{Introduction and Objectives}

To improve the already high availability rate of aircraft engine, health monitoring is developed. 
This process consists in ground based monitoring of numerous measurements made
on the engine and its environment during the aircraft operation. 

One of the goals of this monitoring is to detect abnormal behaviour of the engine that are
early signs of potential failures. 
This detection is done through the analysis of data coming from sensors embedded in the engine.
 Flight after flight, measurements, such as exhausted gas temperature (EGT) and high
pressure (HP) core speed (N2) form a time series.

On one hand, missing such an early sign can lead to operational events
such as in flight shut down. Such operational events can
cause high maintenance costs. On the other
hand, a false alarm (detecting an anomaly when the engine is behaving
normally) can have also costly consequences such as useless engine removal procedure. 

Thus to minimize false alarm, each potential anomaly has to be confirmed
by a human operator. He is then in charge of the identification of the origin of the anomaly.
The long term goal of engine manufacturers is to help companies to minimize
their maintenance costs by giving maintenance recommendations as accurate as
possible. 
Human operators have a very important role in the current industrial process: 
the goal is to help them make improved decisions thanks to a grey box classifier, mainly because
the complexity of the problem seems to prevent any fully automated decision
making. 

The methodology introduced in this paper aims at helping human operators by
leveraging expert knowledge and relying on feature selection to keep only a
small number of binary indicators.  

\subsection{Health monitoring}

Monitoring is strongly based on experts knowledge and field experience. 
Faults and early signs of failures are identified from suitable measurements 
associated to adapted computational transformations of the data. We refer the
reader to e.g. \cite{rabenoroinstants} for examples of the types of
measurements and transformations that can be used in practice.

One of the main difficulties faced by the experts consists in removing from the
measurements any dependency from the flight context.
This normalization process is extremely important as it allows one to assume stationarity of the
residual signal and therefore to leverage change detection methods. In
practice, experts build some anomaly score from those stationarity hypotheses
and when the score passes a limit, the corresponding early sign of failure is
signalled to the human operator. See \cite{come2010aircraft},
\cite{flandrois2009expertise} and \cite{lacaille2009maturation} for some
examples.

One of the problems induced by this general approach is that experts are
generally specialized on a particular subsystem, thus each anomaly score is
mainly focused on a particular subsystem despite the need of a diagnostic of
the whole system. This task is done by human operator who collects all
available information about the desired engine. One of the benefits of the
proposed methodology is its ability to handle binary indicators coming from
all subsystems in an integrated way, as explained in the next section. 

\section{Methodology}\label{sec:method}
The suggested methodology is based on
the selection and combination of a large number of binary indicators. 
While this idea is not entirely new (see e.g.,
\cite{fleuret-2004,hegedus2011methodology}), the methodology proposed here has
some specific aspects. Rather than relying on very basic detectors as in
\cite{fleuret-2004} or on fixed high level expertly designed ones as in
\cite{hegedus2011methodology}, our method takes an intermediate approach: it
varies the parameters of a set of expertly designed parametric indicators. In
addition, it aims at providing an interpretable model. This section details the
proposed procedure. 

\subsection{Expert knowledge}\label{sec:expert-knowledge}
\begin{figure}[htbp]
\centering
\includegraphics[width=0.7\linewidth]{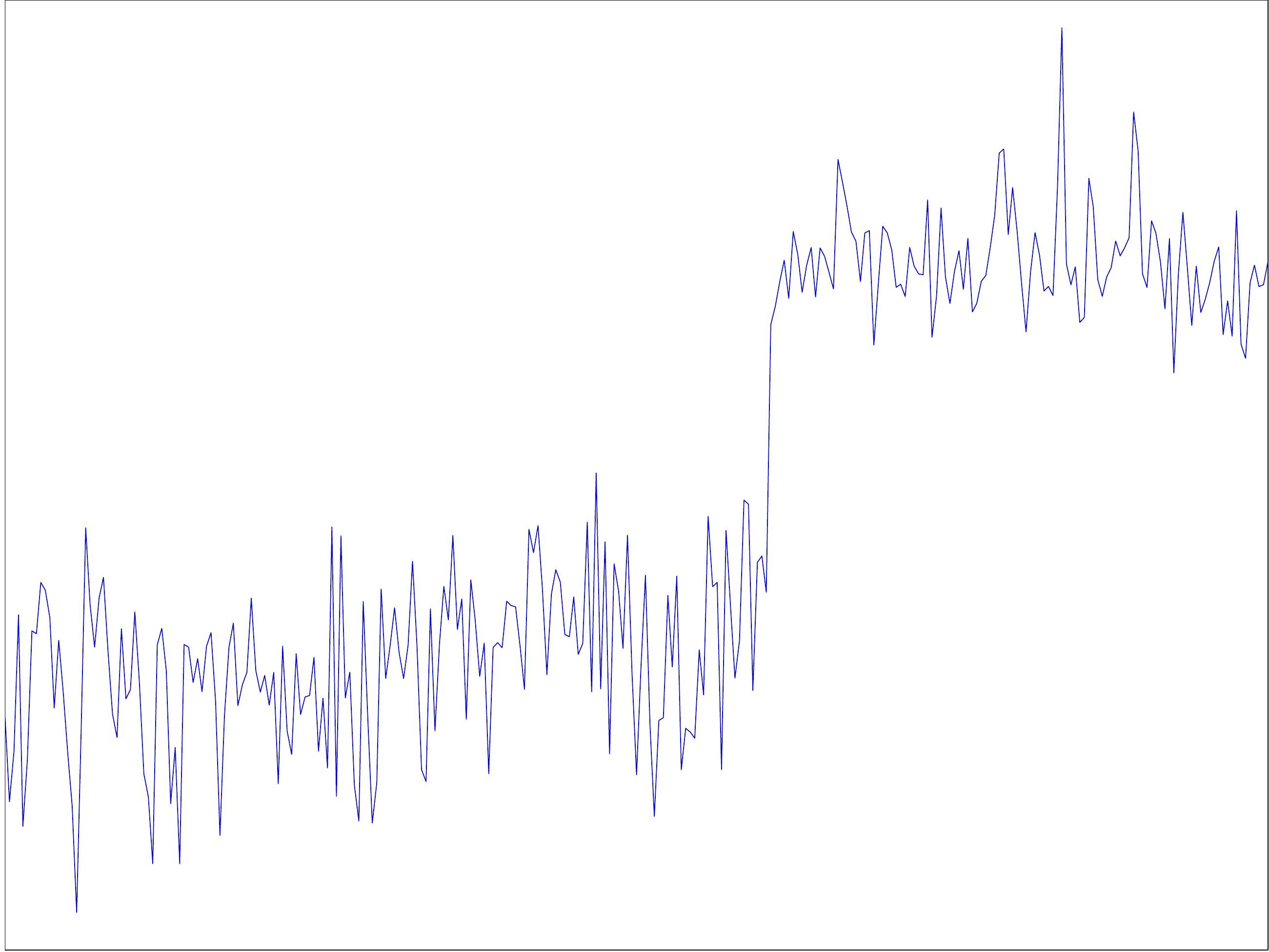}
\caption{Mean shift in a real world time series.}
\label{fig:meanacars}
\end{figure}

This article focuses on change detection
based on statistical techniques \cite{basseville1995detection}. In many
contexts, experts can generally describe more or less explicitly the type of
change they are expecting for some specific (early signs of) anomalies. In the
proposed application context, one can observe for instance a mean shift as in
Figure \ref{fig:meanacars}. 

More generally, experts can describe aggregation and transformation
techniques of raw signals that lead to quantities which should behave in a
``reasonable manner'' under normal circumstances. This can in general be
summarized by computing a distance between the actual quantities and their
expected values. 

\subsection{Exploring parameters space}
In practice however, experts can seldom provide detailed parameter settings
for the aggregation and transformation techniques they recommend. Fixing the
threshold above which a distance from the ``reasonable values'' becomes
critical is also difficult. 

Let us consider for illustration purposes that the expert recommends to look
for shifts in mean of a certain quantity as early signs of a specific anomaly
(as in Figure \ref{fig:meanacars}). If the expert has no strong prior on the distribution of the quantity, a
usual test would be the Mann-Whitney U test. 

Then, one has to assess the scale of the shift. 
The expert has to specify the length of time windows (that defines the scale at which the shift may appear) of the two compared populations.
In most cases, the experts can only give a rough idea of the
scale. 

Given the choice of the test, of its scale and of a change point, 
to take a decision, one has to choose a level to which the $p$-value
will be compared. 

So all in one, looking for a mean shift can be done by choosing at least
three parameters: the type of the test, the scale at which the shift can occur
and the level of the test. The methodology consists in
considering (a subset of) all possible combinations of parameters compatible
with expert knowledge to generate binary indicators. 
 This is a form of indirect grid search procedures for meta-parameter optimisation. 

\subsection{Confirmation indicators}\label{sec:conf-indic}
Finally, aircraft engines are extremely reliable, a
fact that increases the difficulty in balancing sensibility and specificity of
anomaly detectors. 
High level confirmation indicators are built from low level tests to alleviate this difficulty. 
For instance, if we monitor the evolution of a quantity on a long period compared to the expected
time scale of anomalies, we can compare the number of times the null
hypothesis of a test has been rejected on the long period with the number of
times it was not rejected, and turn this into a binary indicator with a
majority rule.

\subsection{Decision}\label{sec:decision}
To summarize, we construct parametric anomaly scores from expert knowledge,
together with acceptable parameter ranges. By exploring those ranges, we
generate numerous (possibly hundreds of) binary indicators. Each indicator can be
linked to an expertly designed score with a specific set of parameters and
thus is supposedly easy to interpret by operators. Notice that while we 
focused in this presentation on temporal data, this framework can be applied
to any data source. 
 
The final decision step consists in classifying these high dimensional binary
vectors in order to further discriminate between seriousness of anomalies
and/or sources (in terms of subsystems of the engine, for instance). 

While including hundreds of indicators is important to give a broad
coverage of the parameters space of the expert scores, 
it seems obvious that some redundancy will appear. 
Moreover reduce the number of indicators will ease the interpretation 
by limiting the quantity of informations transmitted to the human operator.
Thus feature selection \cite{guyon2003introduction} is appropriate.
Unlike \cite{hegedus2011methodology} who choose features by random
projection, the proposed methodology favours interpretable solutions, even at
the expense of the classification accuracy: the goal is to help the human
operator, not to replace her/him. 
Among the possible solutions, we choose to use the Mutual information based
technique Minimum Redundancy Maximum Relevance (mRMR, \cite{peng2005feature})
which was reported to give excellent results on high dimensional data (see
also \cite{fleuret-2004} for another possible choice).

In the considered context, black box modelling is not acceptable, so while
numerous classification algorithms are available (see
e.g. \cite{kotsiantis2007supervised}), we shall focus on interpretable ones. 
Random Forests \cite{breiman2001random} are chosen as the reference
method as they are very adapted to high dimensional
data and known to be robust and to provide state-of-the-art
classification performances. While they are
not as interpretable as their ancestors CART \cite{breiman1984classification},
they provide at least variable importance measures that can be used to
identify the most important indicators.

Another classification algorithm used in this paper is Naive Bayes classifier
\cite{koller2009probabilistic} which is also appropriate for high dimensional
data. They are known to provide good results despite the strong assumption of
the independence of features given the class. In addition, decisions taken by
a Naive Bayes classifier are very easy to understand thanks to the estimation
of the conditional probabilities of the feature in each class. Those
quantities can be shown to the human operator as references. 

\section{Experiments}\label{sec:results}
The proposed methodology 
is evaluated on simulated data which have been modelled based on
real world data such as the ones shown on Figure 
\ref{fig:meanacars}. 

\subsection{Simulated data}
We consider univariate time series of variable length in which three types of
shifts can happen: the mean shift described in Section
\ref{sec:expert-knowledge}, together with a variance and a trend shift described below. Two
data sets are generated, $A$ and $B$. 

In both cases, it is assumed that expert based normalization has been
performed. Therefore when no shift in the data distribution occurs, we observe 
a stationary random noise modelled by the standard Gaussian distribution, that
is $n$ random variables $X_1,\ldots, X_n$ independent and identically
distributed according to $\mathcal{N}(\mu=0,\sigma^2=1)$. 
Signals have a length chosen uniformly at random between 100 and 200 observations 

The three types of shift are :
\begin{enumerate}
\item a variance shift: in this case,
  observations are distributed according to
  $\mathcal{N}(\mu=0,\sigma^2)$ with $\sigma^2=1$ before the change point and $\sigma$ chosen uniformly
  at random in $[1.01, 5]$ after the change point;
\item a mean shift: in this case,
  observations are distributed according to
  $\mathcal{N}(\mu,\sigma^2=1)$ with $\mu=0$ before the change point and $\mu$ chosen uniformly
  at random in $[1.01, 5]$ after the change point in set $A$. Set $B$ is more
  difficult on this aspect as $\mu$ after the change point is chosen uniformly
  at random in $[0.505, 2.5]$;
\item a trend shift: in this case, observations are distributed according to
  $\mathcal{N}(\mu,\sigma^2=1)$ with $\mu=0$ before the change point and $\mu$
  increasing linearly from $0$ from the change point with a slope of chosen uniformly
  at random in $[0.02,3]$.
\end{enumerate}
Assume that the signal contains $n$ observations, then the change point is
chosen uniformly at random between the $\frac{2n}{10}$-th observation and the
$\frac{8n}{10}$-th observation. 
We generate according to this procedure two balanced data set with 6000
observations corresponding to 3000 observations with no anomaly, and 1000
observations for each of the three types of anomalies. 

\subsection{Indicators}
As explained in Section \ref{sec:method}, binary indicators are
constructed from expert knowledge by varying parameters, including scale. 
In the present context, sliding windows are used: for each
position of the window, a classical statistical test is conducted to decide
whether a shift in the signal occurs at the center of the window. 

The ``expert'' designed tests are for these indicators are
 the Mann-Whitney-Wilcoxon U test (non parametric test for shift in
  mean), the two sample Kolmogorov-Smirnov test (non parametric test for
  differences in distributions), the F-test for equality of variance (parametric test based on a Gaussian
  hypothesis). 

The direct parameters of those tests are the size of the window which defines
the two samples (30, 50, and $\min(n-2,100)$ where $n$ is the signal length) and
the level of significance of the test (0.005, 0.1 and 0.5). Notice that those
tests do not include a slope shift detection. 

Then, confirmatory indicators are generated, as explained in Section
\ref{sec:conf-indic}:
\begin{enumerate}
\item for each underlying test, the derived binary indicator takes the value
  one if on $\beta\times m$ windows out of $m$, the test detects a change. 
  Parameters are the test itself with its parameters, the value of $\beta$
  (we considered 0.1, 0.3 and 0.5) and the number of observations in common
  between two consecutive windows (the length of the window minus 1, 5 or
  10);
\item for each underlying test, the derived binary indicator takes the value
  one if on $\beta\times m$ consecutive windows out of $m$, the test detects a
  change (same parameters);
\item for each underlying test, the derived binary indicator takes the value
  one if there are 5 consecutive windows such that the test detects a change
  on at least $k$ of these 5 consecutive windows (similar parameters where $\beta$ is replaced by $k$).
\end{enumerate}
In addition, based on expert recommendations, all those indicators are applied
both to the original signal and to a smoothed signal (using a simple moving
average of 5 observations). 

\subsection{Performance analysis}
Each data set is split in a balanced way into a learning set with 1000 signals
and a test set with 5000 signals. We report the global classification accuracy
(the classification accuracy is the percentage of correct predictions,
regardless of the class) on the learning set to monitor possible over
fitting. The performances of the methodology are evaluated on 10 balanced
subsets of size 500 from the 5000 signals' test set. This allows to evaluate
both the average performances and their variabilities. For the Random Forest, we
also report the out-of-bag (oob) estimate of the classification accuracy (this
is a byproduct of the bootstrap procedure used to construct the forest, see
\cite{breiman2001random}). Finally, we use confusion matrices and class
specific accuracy to gain more insights on the results when needed. 

\subsection{Performances with all indicators}
As indicators are expertly designed and should cover the useful
parameter range of the tests, it is assumed that the best classification
performances should be obtained when using all of them, up to the effects of
the curse of dimensionality. 

\begin{table}[htbp]
  \caption{Classification accuracy using 810 binary
    indicators. For the test set, we report the average classification
    accuracy and its standard deviation between parenthesis.}
  \label{tab:fullRF}
  \centering
		\vskip 0.15in
		\begin{small}
  \begin{tabular}{lccc}
\multicolumn{3}{c}{Random Forest}\\
Data & Training acc. & OOB acc. & Test average acc. \\\hline
$A$ & 0.9770 & 0.9228 &0.9352 (0.0100) \\
$B$ &  0.9709 & 0.9118 & 0.9226 (0.0108) \\\hline
& & & \\
\multicolumn{3}{c}{Naive Bayesian Classifier}\\
Data & Training acc. & - & Test average acc. \\\hline
$A$ &0.9228   & - &0.8687 (0.0099) \\
$B$ & 0.8978 & - &0.8632 (0.0160) \\\hline
  \end{tabular}
	\end{small}
\medskip

\end{table}

Table \ref{tab:fullRF} reports the global classification accuracy of the
Random Forest and Naive Bayes classifier, using all the indicators. As expected, Random Forests suffer
neither from the curse of dimensionality nor from strong over fitting (the
test set performances are close to the learning set ones). For the Naive Bayes
classifier, those performances are significantly lower than the one obtained
by the Random Forest. As shown by the confusion matrix on Table
\ref{tab:conf:NBNsetAtest}, the classification errors are not concentrated on
one class (even if the errors are not perfectly balanced). This tends to
confirm that the indicators are adequate to the task (this was already obvious
from the Random Forest).

\begin{table}[h]
\caption{Data set $A$: confusion matrix with all indicators for Naive Bayes
  classifier on the full test set.}
\label{tab:conf:NBNsetAtest}
\vskip 0.15in
\centering

\begin{tabular}{c|ccccc}
&0&1&2&3&total\\\hline
0&2267&162&32&39&2500\\
1&118&671&36&4&829\\
2&26&4&708&91&829\\
3&46&7&76&700&829\\\hline
\end{tabular}

\vskip -0.1in
\end{table}

\subsection{Feature selection}
While satisfactory results are obtained,  
it would be unrealistic to ask to an
operator to review 810 binary values to understand why the classifier favours
one class rather than the others. Thus a feature selection should be appropriate.

As explained in Section \ref{sec:decision}, the feature selection relies on
the mRMR ranking procedure. A forward approach is used to evaluate how many
indicators are needed to achieve acceptable predictive performances. Notice
that in the forward approach, indicators are added in the order given by mRMR
and then are never removed. As mRMR takes into account redundancy between the
indicators, this should not be a major issue. Then for each number of
indicators, a Random Forest and a Naive Bayes classifier are constructed and
evaluated. 

\begin{figure}[htbp]
\centerline{\includegraphics[width=0.9\linewidth]{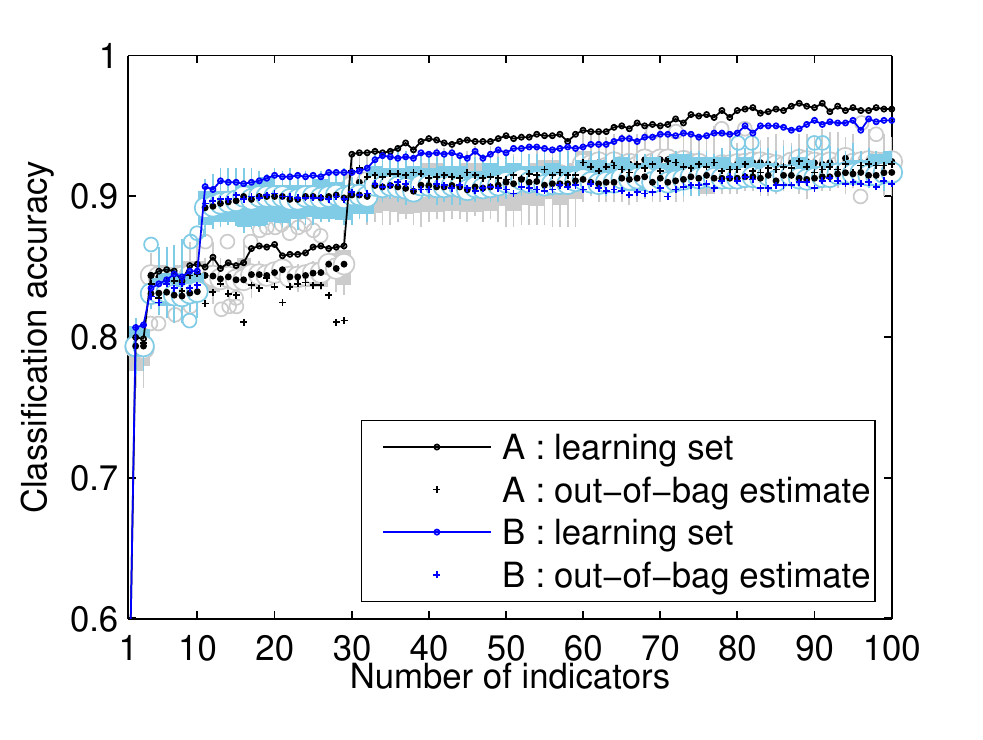}}
\caption{\textbf{Data sets  $A$ (black) and $B$ (blue) Random Forest}: classification accuracy on learning set (circle) as a function of the
  number of indicators. A boxplot gives the classification accuracies on the
  test subsets, summarized by its median (black dot inside a white
  circle). The estimation of those accuracies by the out-of-bag (oob) bootstrap
  estimate is shown by the crosses.}
\label{fig:rfboxsetA}
\end{figure}

\begin{figure}[htbp]
\centering
\includegraphics[width=0.9\linewidth]{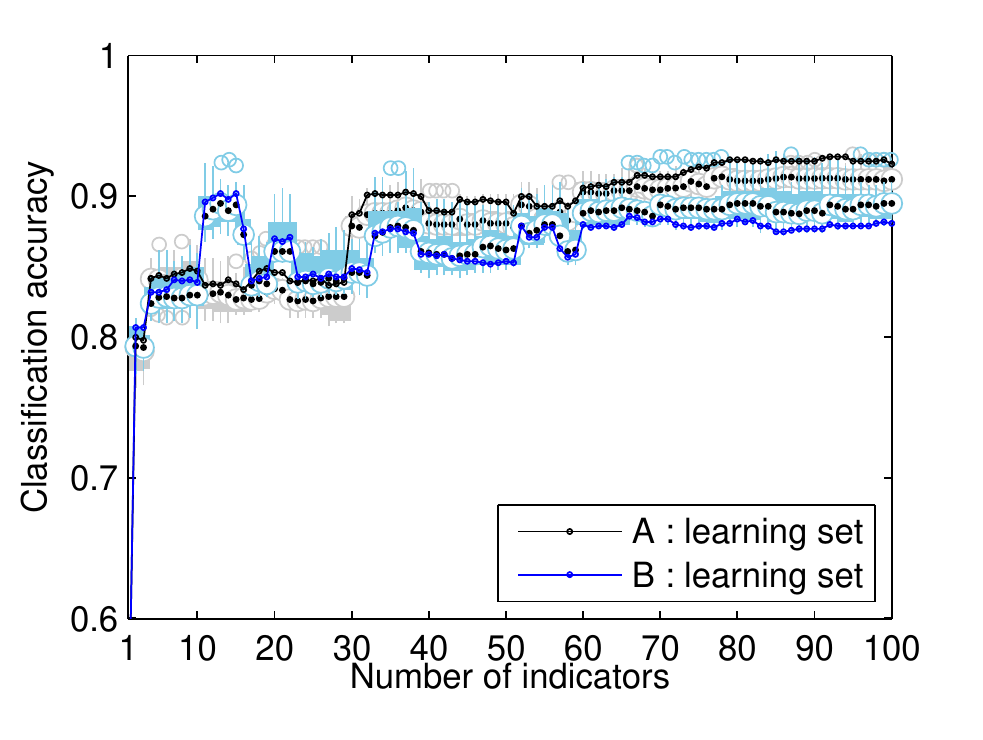}
\caption{\textbf{Data sets  $A$ (black) and $B$ (blue)  Naive Bayes classifier}: classification accuracy on learning set (circle) as a function of the
  number of indicators. A boxplot gives the classification accuracies on the
  test subsets, summarized by its median (black dot inside a white
  circle). }
\label{fig:nbnboxsetA}
\end{figure}

Figures \ref{fig:rfboxsetA} and \ref{fig:nbnboxsetA} 
 summarize the results for the 100 first indicators for sets $A$ and $B$. The
classification accuracy of the Random Forest increases almost monotonously
with the number of indicators, but after roughly 25 to 30 indicators
(depending on the data set), performances on the test set tend to stagnate
. In practice, this means that the proposed procedure can be used to
select the relevant indicators implementing this way an automatic tuning
procedure for the parameters of the expertly designed scores. 

Results for the Naive Bayes classifier are slightly more complex in the case
of the second data set, but they confirm that indicator selection is
possible. Notice that the
learning set performances of the Naive Bayes classifier are almost identical
to its test set performances (which exhibit almost no variability over the
slices of the full test set). This is natural because the classifier is based
on the estimation of the probability of observing a 1 value
\emph{independently} for each indicator, conditionally on the class. The
learning set contains at least 250 observations for each class, leading to a
very accurate estimation of those probabilities and thus to very stable
decisions. In practice one can therefore select the optimal number of
indicators using the learning set performances, without the need of a
cross-validation procedure.

\begin{figure}[htbp]
\centering
\includegraphics[width=0.9\linewidth]{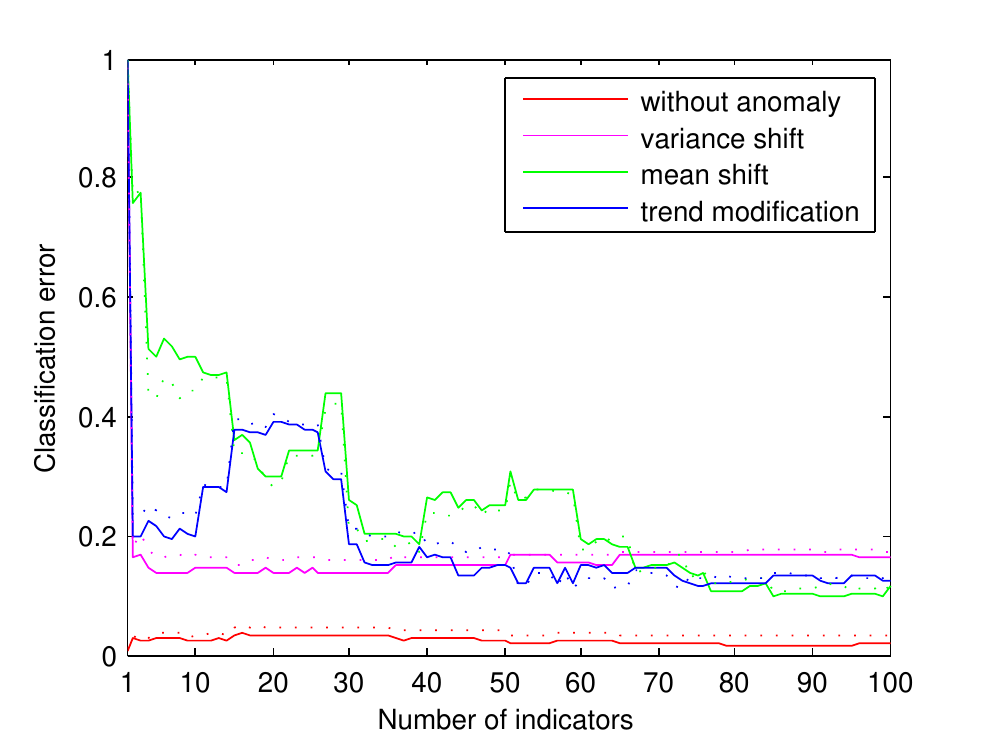}
\caption{\textbf{Data set $A$ Naive Bayes classifier}: classification error
  for each class on the training set (solid lines) and on the test set (dotted
  lines, average accuracies only).}
\label{fig:nbperclassA}
\end{figure}

It should be noted that significant jumps in performances can be observed in
all cases. This might be an indication that the ordering provided by the mRMR
procedure is not optimal. A possible solution to reach better indicator
subsets would be to use a wrapper approach, leveraging the computational
efficiency of both Random Forest and Naive Bayes construction. Meanwhile
Figure \ref{fig:nbperclassA} shows in more detail this phenomenon by
displaying the classification error class by class, as a function of the
number of indicators, in the case of data set $A$. The figure shows the
difficulty of discerning between mean shift and trend shift (for the latter,
no specific test has been included, on purpose). But as the strong decrease in
classification error when the 30-th indicator is added concerns both classes
(mean shift and trend shift), the ordering provided by mRMR could be
questioned. 

\subsection{Indicator selection}
Based on results shown on Figure \ref{fig:nbnboxsetA},  
 one can select an optimal number of binary indicators,
while enforcing a reasonable limit on this number to avoid flooding the human
operator with too many results. For instance Table \ref{tab:NB:selected} gives
the classification accuracy of the Naive Bayes classifier using the optimal
number of binary indicators between 1 and 20. 

\begin{table}[htbp]
\caption{Classification accuracy of the Naive Bayesian network using the
optimal number binary indicators between 1 and 20. For the test set, we report the average
classification accuracy and its standard deviation between parenthesis.}
\label{tab:NB:selected}
\vskip 0.15in
\centering
\begin{small}
\begin{tabular}{lcccc}
Data & Training acc. & Test avg acc. & \# indicators\\\hline
$A$ & 0.8487 & 0.8448 (0.0134) & 9\\
$B$ &0.9018 & 0.8935 (0.0130) & 13\\ \hline
\end{tabular}
\end{small}
\end{table}

While the performances are not as good as the ones of the Random Forest, they
are acceptable : the selected indicators can be shown to the human operator together
with the estimated probabilities of getting a positive result from each
indicator, conditionally on each class, shown on Table
\ref{tab:tenbest:A}. For instance here the first selected indicator,
$confu(2,3)$, is a confirmation indicator for the U test. It is positive when
there are 2 windows out of 3 consecutive ones on which a U test was
positive. The Naive Bayes classifier uses the estimated probabilities to reach
a decision: here the indicator is very unlikely to be positive if there is no
change or if the change is a variance shift. On the contrary, it is very
likely to be positive when there is a mean or a trend shift. While the table
does not ``explain'' the decisions made by the Naive Bayes classifier, it
gives easily interpretable hints to the human operator.  

\begin{table}[htbp]
\caption{The 9 best indicators according to mRMR for data set
$A$. Confu(k,n) corresponds to a positive Mann-Whitney-Wilcoxon U test on
k windows out of n consecutive ones. Conff(k,n) is the same thing for the
F-test. Ratef($\alpha$) corresponds to a positive F-test on $\alpha\times m$
windows out of $m$. Lseqf($\alpha$) corresponds to a positive F-test on
$\alpha\times m$ consecutive windows out of $m$. Lsequ($\alpha$) is the same for a U test.
Detailed parameters of the indicators have been omitted for brevity.}
\label{tab:tenbest:A}
\vskip 0.15in
\centering
\begin{small}
\begin{tabular}{ccccc}
type of indicator& no change &variance&mean&trend\\\hline

confu(2,3)&0.0103&0.011&0.971&0.939\\
F test&0.0206&0.83&0.742&0.779\\
U test&0.02&0.022&0.968&0.941\\
lseqf(0.3)&0.0053&0.571&0.336&0.023\\
confu(4,5)&0.0343&0.03&0.986&0.959\\
confu(3,5)&0.0013&0.001&0.923&0.899\\
confu(2,3)&0.0673&0.06&0.992&0.967\\
F test&0.042&0.853&0.793&0.813\\
KS test&0.0113&0.259&0.961&0.923\\\hline
\end{tabular}

\end{small}
\vskip -0.1in
\end{table}

\subsection{Role of confirmation indicators}

In table \ref{tab:tenbest:A}, one can see several confirmation indicators. 
In this section, we illustrate their impacts on results.
We use a new simulated data set $C$. 
Its random noise is based on a Student distribution ($3$ degrees of freedom) but with a random variance.
That is $n$ random variables $Y_1,\ldots, Y_n$  independent and identically distributed according to $ \mathcal{T}(3)$.
$X_1,\ldots, X_n$ are such that  
$  X_i \sim a Y_i$  where $a$ is chosen uniformly at random in $[0.5,3]$.
The same three types of shift are used. But, the mean shift added is chosen uniformly at random in $[0.3, 5]$ and 
the variance shift is added chosen at random in $[1.05, 5]$.

For $C$, new tests are added : the Student-test with equal variance and unequal variance (test for shift in mean),
a slope shift detection and a slope change test. 2565 indicators are then obtained.

\begin{figure}[htbp]
\centerline{\includegraphics[width=0.9\linewidth]{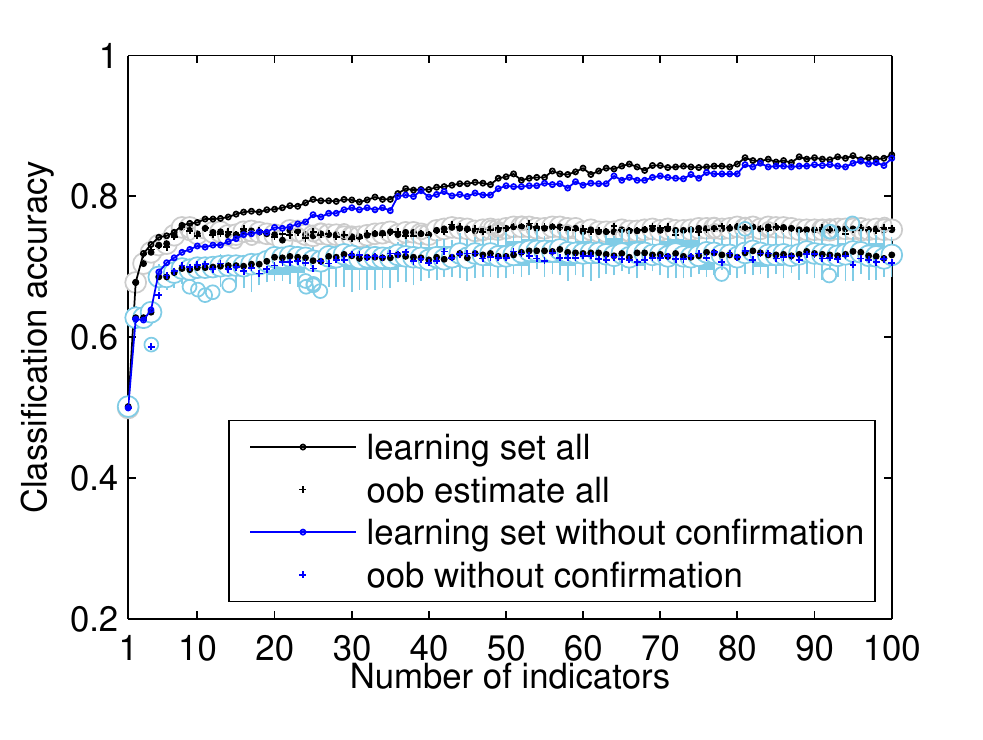}}
\caption{\textbf{Data set $C$ (black) and $Cm$(blue) Random Forest}, see Figure \ref{fig:rfboxsetA}
  for details. $Cm$ is the same data as $C$ but no confirmation indicators are used.}
\label{fig:rfboxsetG}
\end{figure}

In table \ref{tab:fullRFconf}, the result obtained with and without confirmation indicators ($Cm$)
is reported. 
In figure \ref{fig:rfboxsetG}, the difference obtained in set $C$ with and without confirmation indicators is given for the first 100 indicators
according to mRMR. The confirmation indicators improve the classification accuracy.

\begin{table}[htbp]
  \caption{Classification accuracy of the Random Forest classifier using all binary
    indicators : 2565 for set $C$, 945 for set $Cm$  . $Cm$ is the same data as $C$ but no confirmation indicators are used.}
  \label{tab:fullRFconf}
  \centering
	\vskip 0.15in
	\begin{small}
  \begin{tabular}{lccc}
Data set & Training set acc. & Test set average acc. \\\hline
$C$ & 0.9629   & 0.7656 ( 0.0161) \\
$Cm$ & 0.95298 & 0.731513 (0.0171) \\\hline
  \end{tabular}
\medskip
\end{small}
\end{table}

\section*{Acknowledgments}
This study is supported by grant from Snecma\footnote{Snecma, Safran Group, is one of the world’s leading manufacturers of aircraft and rocket engines, see
 \url{http://www.snecma.com/} for details.}.

\section{Conclusion and perspectives}
This paper proposes a general methodology that combines expert knowledge with
feature selection and automatic classification to design accurate anomaly
detector and classifier. Feature selection allows to reduce the
number of useful indicators to a humanly manageable number. This allows a
human operator to understand at least partially how a decision is reached by
an automatic classifier. This is favoured by the choice of the indicators which
are based on expert knowledge. A very interesting byproduct of the
methodology is that it can work on very different original data as long as
expert decision can be modelled by a set of parametric anomaly scores. This was
illustrated by working on signals of different lengths. 

The methodology has been shown sound  using simulated data. Using a reference
high performance classifier, Random Forests, the indicator generation
technique covers sufficiently the parameters space to obtain high
classification rate. Then, the feature selection mechanism leads to a reduced number of indicators 
with good predictive performances when paired with a
simpler classifier, the Naive Bayes classifier. As shown in the experiments,
the class conditional probabilities of obtaining a positive value for those
indicators provide interesting insights on the way the Naive Bayes classifier
takes a decision. 

In order to justify the cost of collecting a sufficiently large real world
labelled data set in our context (engine health monitoring), additional
experiments are needed. In particular, multivariate data must be studied in
order to simulate the case of a complex system made of numerous
sub-systems. This will naturally lead to more complex anomaly models. We also
observed possible limitations of the feature selection strategy used here as
the performances displayed abrupt changes during the forward procedure. More
computationally demanding solutions, namely wrapper ones, will be studied to
confirm this point. 

It is also important to notice that the classification accuracy is not the
best way of evaluating the performances of a classifier in the health
monitoring context. Firstly, health monitoring involves intrinsically a strong
class imbalance \cite{japkowicz2002class}. Secondly, health monitoring is a
cost sensitive area because of the strong impact on airline profit of an
unscheduled maintenance. It is therefore important to take into account
specific asymmetric misclassification cost to get a proper performance
evaluation. For example, results have shown the role played by confirmation indicators,
 they are designed to limit false alarm rate.

\bibliography{Biblio}

\begin{thebibliography}{}

\bibitem[Basseville \& Nikiforov, 1995][Basseville and
  Nikiforov][1995]{basseville1995detection}
Basseville, M., \& Nikiforov, I.~V. (1995).
\newblock Detection of abrupt changes: theory and applications.
\newblock {\em Journal of the Royal Statistical Society-Series A Statistics in
  Society}, {\em 158}, 185.

\bibitem[Breiman, 2001][Breiman][2001]{breiman2001random}
Breiman, L. (2001).
\newblock Random forests.
\newblock {\em Machine learning}, {\em 45}, 5--32.

\bibitem[Breiman et~al.\/, 1984][Breiman
  et~al.\/][1984]{breiman1984classification}
Breiman, L., Friedman, J.~H., Olshen, R.~A., \& Stone, C.~J. (1984).
\newblock Classification and regression trees. wadsworth \& brooks.
\newblock {\em Monterey, CA}.

\bibitem[Chandola et~al.\/, 2009][Chandola et~al.\/][2009]{chandola2009anomaly}
Chandola, V., Banerjee, A., \& Kumar, V. (2009).
\newblock Anomaly detection: A survey.
\newblock {\em ACM Computing Surveys (CSUR)}, {\em 41}, 15.

\bibitem[C{\^o}me et~al.\/, 2010][C{\^o}me et~al.\/][2010]{come2010aircraft}
C{\^o}me, E., Cottrell, M., Verleysen, M., \& Lacaille, J. (2010).
\newblock Aircraft engine health monitoring using self-organizing maps.
\newblock In {\em Advances in data mining. applications and theoretical
  aspects},  405--417. Springer.

\bibitem[Flandrois et~al.\/, 2009][Flandrois
  et~al.\/][2009]{flandrois2009expertise}
Flandrois, X., Lacaille, J., Masse, J.-R., \& Ausloos, A. (2009).
\newblock Expertise transfer and automatic failure classification for the
  engine start capability system.
\newblock {\em AIAA Infotech, Seattle, WA}.

\bibitem[Fleuret, 2004][Fleuret][2004]{fleuret-2004}
Fleuret, F. (2004).
\newblock Fast binary feature selection with conditional mutual information.
\newblock {\em Journal of Machine Learning Research (JMLR)}, {\em 5},
  1531--1555.

\bibitem[Guyon \& Elisseeff, 2003][Guyon and
  Elisseeff][2003]{guyon2003introduction}
Guyon, I., \& Elisseeff, A. (2003).
\newblock An introduction to variable and feature selection.
\newblock {\em The Journal of Machine Learning Research}, {\em 3}, 1157--1182.

\bibitem[Hegedus et~al.\/, 2011][Hegedus
  et~al.\/][2011]{hegedus2011methodology}
Hegedus, J., Miche, Y., Ilin, A., \& Lendasse, A. (2011).
\newblock Methodology for behavioral-based malware analysis and detection using
  random projections and k-nearest neighbors classifiers.
\newblock {\em Computational Intelligence and Security (CIS), 2011 Seventh
  International Conference on} (pp.\/ 1016--1023).

\bibitem[Japkowicz \& Stephen, 2002][Japkowicz and
  Stephen][2002]{japkowicz2002class}
Japkowicz, N., \& Stephen, S. (2002).
\newblock The class imbalance problem: A systematic study.
\newblock {\em Intelligent data analysis}, {\em 6}, 429--449.

\bibitem[Koller \& Friedman, 2009][Koller and
  Friedman][2009]{koller2009probabilistic}
Koller, D., \& Friedman, N. (2009).
\newblock {\em Probabilistic graphical models: principles and techniques}.
\newblock The MIT Press.

\bibitem[Kotsiantis et~al.\/, 2007][Kotsiantis
  et~al.\/][2007]{kotsiantis2007supervised}
Kotsiantis, S.~B., Zaharakis, I., \& Pintelas, P. (2007).
\newblock Supervised machine learning: A review of classification techniques.

\bibitem[Lacaille, 2009][Lacaille][2009]{lacaille2009maturation}
Lacaille, J. (2009).
\newblock A maturation environment to develop and manage health monitoring
  algorithms.
\newblock {\em PHM, San Diego, CA}.

\bibitem[Peng et~al.\/, 2005][Peng et~al.\/][2005]{peng2005feature}
Peng, H., Long, F., \& Ding, C. (2005).
\newblock Feature selection based on mutual information criteria of
  max-dependency, max-relevance, and min-redundancy.
\newblock {\em Pattern Analysis and Machine Intelligence, IEEE Transactions
  on}, {\em 27}, 1226--1238.

\bibitem[Rabenoro \& Lacaille, 2013][Rabenoro and
  Lacaille][2013]{rabenoroinstants}
Rabenoro, T., \& Lacaille, J. (2013).
\newblock Instants extraction for aircraft engine monitoring.
\newblock {\em AIAA Infotech@Aerospace}.

\end{thebibliography}
\bibliographystyle{mlapa}

\end{document}